# Regression Conformal Prediction with Nearest Neighbours

**Harris Papadopoulos**                               H.PAPADOPOULOS@FREDERICK.AC.CY
*Computer Science and Engineering Department*
*Frederick University*
*7 Y. Frederickou St., Palouriotisa*
*Nicosia 1036, Cyprus*

**Vladimir Vovk**                                          VOVK@CS.RHUL.AC.UK
**Alex Gammerman**                                      ALEX@CS.RHUL.AC.UK
*Computer Learning Research Centre*
*Department of Computer Science*
*Royal Holloway, University of London*
*Egham, Surrey TW20 0EX, UK*

## Abstract

In this paper we apply Conformal Prediction (CP) to the $k$-Nearest Neighbours Regression ($k$-NNR) algorithm and propose ways of extending the typical nonconformity measure used for regression so far. Unlike traditional regression methods which produce point predictions, Conformal Predictors output predictive regions that satisfy a given confidence level. The regions produced by any Conformal Predictor are automatically valid, however their tightness and therefore usefulness depends on the nonconformity measure used by each CP. In effect a nonconformity measure evaluates how strange a given example is compared to a set of other examples based on some traditional machine learning algorithm. We define six novel nonconformity measures based on the $k$-Nearest Neighbours Regression algorithm and develop the corresponding CPs following both the original (transductive) and the inductive CP approaches. A comparison of the predictive regions produced by our measures with those of the typical regression measure suggests that a major improvement in terms of predictive region tightness is achieved by the new measures.

## 1. Introduction

A drawback of traditional machine learning algorithms is that they do not associate their predictions with confidence information, instead they only output *simple predictions*. However, some kind of confidence information about predictions is of paramount importance in many risk-sensitive applications such as those used for medical diagnosis (Holst, Ohlsson, Peterson, & Edenbrandt, 1998).

Of course some machine learning theories that produce confidence information do exist. One can apply the theory of Probably Approximately Correct learning (PAC theory, Valiant, 1984) to an algorithm in order to obtain upper bounds on the probability of its error with respect to some confidence level. The bounds produced by PAC theory though, will be very weak unless the data set to which the algorithm is being applied is particularly clean, which is rarely the case. Nouretdinov, Vovk, Vyugin, and Gammerman (2001b) demonstrated the crudeness of PAC bounds by applying one of the best bounds, by Littlestone and Warmuth (Cristianini & Shawe-Taylor, 2000, Thm. 4.25, 6.8), to the USPS data set.

Another way of obtaining confidence information is by using the Bayesian framework for producing algorithms that complement individual predictions with probabilistic measures of their quality. In order to apply the Bayesian framework however, one is required to have some prior knowledge about the distribution generating the data. When the correct prior is known, Bayesian methods provide optimal decisions. For real world data sets though, as the required knowledge is not available, one has to assume the existence of an arbitrarily chosen prior. In this case, since the assumed prior may be incorrect, the resulting confidence levels may also be "incorrect"; for example the predictive regions output for the 95% confidence level may contain the true label in much less than 95% of the cases. This signifies a major failure as we would expect confidence levels to bound the percentage of expected errors. An experimental demonstration of this negative aspect of Bayesian methods in the case of regression is given in Section 8, while a detailed experimental examination for both classification and regression was performed by Melluish, Saunders, Nouretdinov, and Vovk (2001).

A different approach to confidence prediction was suggested by Gammerman, Vapnik, and Vovk (1998) (and later greatly improved by Saunders, Gammerman, & Vovk, 1999), who proposed what we call in this paper "Conformal Prediction" (CP). A thorough analysis of CP was given by Vovk, Gammerman, and Shafer (2005), while an overview was presented by Gammerman and Vovk (2007). Conformal Predictors are built on top of traditional machine learning algorithms and accompany each of their predictions with valid measures of confidence. Unlike Bayesian methods, CPs do not require any further assumptions about the distribution of the data, other than that the data are independently and identically distributed (i.i.d.); although this is still a strong assumption, it is almost universally accepted in machine learning. Even if the traditional algorithm on which a CP is based makes some extra assumptions that are not true for a particular data set, the validity of the predictive regions produced by the CP will not be affected. The resulting predictive regions might be uninteresting, but they will still be valid, as opposed to the misleading regions produced by Bayesian methods. Furthermore, in contrast to PAC methods, the confidence measures they produce are useful in practice. Different variants of CPs have been developed based on Support Vector Machines (Saunders et al., 1999; Saunders, Gammerman, & Vovk, 2000), Ridge Regression (Nouretdinov, Melluish, & Vovk, 2001a; Papadopoulos, Proedrou, Vovk, & Gammerman, 2002a), $k$-Nearest Neighbours for classification (Proedrou, Nouretdinov, Vovk, & Gammerman, 2002; Papadopoulos, Vovk, & Gammerman, 2002b) and Neural Networks (Papadopoulos, Vovk, & Gammerman, 2007), all of which have been shown to give reliable and high quality confidence measures. Moreover, CP has been applied successfully to many problems such as the early detection of ovarian cancer (Gammerman et al., 2009), the classification of leukaemia subtypes (Bellotti, Luo, Gammerman, Delft, & Saha, 2005), the diagnosis of acute abdominal pain (Papadopoulos, Gammerman, & Vovk, 2009a), the prediction of plant promoters (Shahmuradov, Solovyev, & Gammerman, 2005), the recognition of hypoxia electroencephalograms (EEGs) (Zhang, Li, Hu, Li, & Luo, 2008), the prediction of network traffic demand (Dashevskiy & Luo, 2008) and the estimation of effort for software projects (Papadopoulos, Papatheocharous, & Andreou, 2009b).

The only drawback of the original CP approach is its relative computational inefficiency. This is due to the transductive nature of the approach, which entails that all computations have to start from scratch for every test example. This renders it unsuitable for application to large data sets. For this reason a modification of the original CP approach, called Inductive Conformal Prediction (ICP), was proposed by Papadopoulos et al. (2002a) for regression and by Papadopoulos et al. (2002b) for classification. As suggested by its name, ICP replaces the transductive inference followed in the original approach with inductive inference. Consequently, ICPs are almost as computationally efficient as their underlying algorithms. This is achieved at the cost of some loss in the quality of the produced confidence measures, but this loss is negligible, especially when the data set in question is large, whereas the improvement in computational efficiency is significant. A computational complexity comparison between the original CP and ICP approaches was performed by Papadopoulos (2008). From now on, in order to differentiate clearly between the original CP and ICP approaches the former will be called Transductive Conformal Prediction (TCP).

In order to apply CP (either TCP or ICP) to a traditional algorithm one has to develop a *nonconformity measure* based on that algorithm. This measure evaluates the difference of a new example from a set (actually a multiset or bag) of old examples. Nonconformity measures are constructed using as basis the traditional algorithm to which CP is being applied, called the *underlying algorithm* of the resulting Conformal Predictor. In effect nonconformity measures assess the degree to which the new example disagrees with the attribute-label relationship of the old examples, according to the underlying algorithm of the CP. It is worth to note that many different nonconformity measures can be constructed for each traditional algorithm and each of those measures defines a different CP. This difference, as we will show in the next section, does not affect the validity of the results produced by the CPs, it only affects their efficiency.

In this paper we are only interested in the problem of regression and we focus on $k$-Nearest Neighbours Regression ($k$-NNR) as underlying algorithm, which is one of the most popular machine learning techniques. The first regression CPs were proposed by Nouretdinov et al. (2001a) following the TCP approach and by Papadopoulos et al. (2002a) following the ICP approach, both based on the Ridge Regression algorithm. As opposed to the conventional point predictions, the output of regression CPs is a predictive region that satisfies a given confidence level.

The typical nonconformity measure used so far in the case of regression is the absolute difference $|y_i - \hat{y}_i|$, between the actual label $y_i$ of the example $i$ and the predicted label $\hat{y}_i$ of the underlying algorithm for that example, given the old examples as training set. Here we propose six extensions to this nonconformity measure for $k$-Nearest Neighbours Regression and develop the corresponding Inductive and Transductive CPs; unfortunately although all six new measures can be used with the ICP approach, only two of them can be used with TCP. Our definitions normalize the standard measure based on the expected accuracy of the underlying algorithm for each example, which makes the width of the resulting predictive regions vary accordingly. As a result, the predictive regions produced by our measures are in general much tighter than those produced by the standard regression measure. This paper extends our previous work (Papadopoulos, Gammerman, & Vovk, 2008) where the $k$-Nearest Neighbours Regression TCP was developed using two normalized nonconformity measures. It is also worth mentioning that one other such nonconformity measure definition was presented by Papadopoulos et al. (2002a) for the Ridge Regression ICP.

The rest of this paper is structured as follows. In the next section we discuss the general idea on which CPs are based. Then in Sections 3 and 4 we describe the $k$-Nearest Neigh-bours Regression TCP and ICP respectively using the typical regression nonconformity measure. In Section 5 we give our new nonconformity measure definitions and explain the rationale behind them. Section 6 analyses further one of our new nonconformity measures and demonstrates that under specific assumptions it gives asymptotically optimal predictive regions. Section 7 details our experimental results with the 3 TCPs and 7 ICPs developed based on the different measures, while Section 8 compares our methods with Gaussian Process Regression (Rasmussen & Williams, 2006), which is one of the most popular Bayesian approaches. Finally, Section 9 gives our conclusions and discusses some possible future directions of this work.

## 2. Conformal Prediction

In this section we briefly describe the idea behind Conformal Prediction; for a more detailed description the interested reader is referred to the book by Vovk et al. (2005). We are given a training set $\{z_1, \ldots, z_l\}$ of examples, where each $z_i \in Z$ is a pair $(x_i, y_i)$; $x_i \in \mathbb{R}^d$ is the vector of attributes for example $i$ and $y_i \in \mathbb{R}$ is the label of that example. We are also given a new unlabeled example $x_{l+1}$ and our task is to state something about our confidence in different values $\tilde{y}$ for the label $y_{l+1}$ of this example. As mentioned in Section 1 our only assumption is that all $(x_i, y_i)$, $i = 1, 2, \ldots$, are generated independently from the same probability distribution.

First let us define the concept of a nonconformity measure. Formally, a nonconformity measure is a family of functions $A_n : Z^{(n-1)} \times Z \to \mathbb{R}$, $n = 1, 2, \ldots$ (where $Z^{(n-1)}$ is the set of all multisets of size $n - 1$), which assign a numerical score

$$\alpha_i = A_n(\{z_1, \ldots, z_{i-1}, z_{i+1}, \ldots, z_n\}, z_i) \tag{1}$$

to each example $z_i$, indicating how different it is from the examples in the multiset

$$\{z_1, \ldots, z_{i-1}, z_{i+1}, \ldots, z_n\}.$$

As mentioned in Section 1 each nonconformity measure is based on some traditional machine learning method, which is called the *underlying algorithm* of the corresponding CP. Given a training set of examples $\{z_1, \ldots, z_{l+1}\}$, each such method creates a prediction rule

$$D_{\{z_1, \ldots, z_{l+1}\}},$$

which maps any unlabeled example $x$ to a label $\hat{y}$. As this prediction rule is based on the examples in the training set, the nonconformity score of an example $z_i \in \{z_1, \ldots, z_{l+1}\}$ is measured as the disagreement between the predicted label

$$\hat{y}_i = D_{\{z_1, \ldots, z_{l+1}\}}(x_i) \tag{2}$$

and the actual label $y_i$ of $z_i$. Alternatively, we can create the prediction rule

$$D_{\{z_1, \ldots, z_{i-1}, z_{i+1}, \ldots, z_{l+1}\}}$$

using all the examples in the set except $z_i$, and measure the disagreement between

$$\hat{y}_i = D_{\{z_1, \ldots, z_{i-1}, z_{i+1}, \ldots, z_{l+1}\}}(x_i) \tag{3}$$

and $y_i$.

Now suppose we are interested in some particular guess $\tilde{y}$ for the label of $x_{l+1}$. Adding this new example $(x_{l+1}, \tilde{y})$ to our known data set $\{(x_1, y_1), \ldots, (x_l, y_l)\}$ gives the extended set

$$\{z_1, \ldots, z_{l+1}\} = \{(x_1, y_1), \ldots, (x_{l+1}, \tilde{y})\}; \tag{4}$$

notice that the only unknown component of this set is the label $\tilde{y}$. We can now use a nonconformity measure $A_{l+1}$ to compute the nonconformity score

$$\alpha_i = A_{l+1}(\{z_1, \ldots, z_{i-1}, z_{i+1}, \ldots, z_{l+1}\}, z_i)$$

of each example $z_i$, $i = 1, \ldots, l+1$ in (4). The nonconformity score $\alpha_{l+1}$ on its own does not really give us any information, it is just a numeric value. However, we can find out how unusual $z_{l+1}$ is according to $A_{l+1}$ by comparing $\alpha_{l+1}$ with all other nonconformity scores. This comparison can be performed with the function

$$p(\tilde{y}) = \frac{\#\{i = 1, \ldots, l+1 : \alpha_i \geq \alpha_{l+1}\}}{l+1} \tag{5}$$

(we leave the dependence of the left-hand side on $z_1, \ldots, z_l, x_{l+1}$ implicit, but it should be always kept in mind). We call the output of this function, which lies between $\frac{1}{l+1}$ and 1, the p-value of $\tilde{y}$, as that is the only part of (4) we were not given. An important property of (5) is that $\forall \delta \in [0, 1]$ and for all probability distributions $P$ on $Z$,

$$P\{\{z_1, \ldots, z_{l+1}\} : p(y_{l+1}) \leq \delta\} \leq \delta; \tag{6}$$

a proof was given by Nouretdinov et al. (2001b). As a result, if the p-value of a given label is below some very low threshold, say 0.05, this would mean that this label is highly unlikely as such sets will only be generated at most 5% of the time by any i.i.d. process.

Assuming we could calculate the p-value of every possible label $\tilde{y}$, as described above, we would be able to exclude all labels that have a p-value under some very low threshold (or *significance level*) $\delta$ and have at most $\delta$ chance of being wrong. Consequently, given a confidence level $1 - \delta$ a regression conformal predictor outputs the set

$$\{\tilde{y} : p(\tilde{y}) > \delta\}, \tag{7}$$

i.e. the set of all labels that have a p-value greater than $\delta$. Of course it would be impossible to explicitly calculate the p-value of every possible label $\tilde{y} \in \mathbb{R}$. In the next section we describe how one can compute the predictive region (7) efficiently for $k$-Nearest Neighbours Regression.

It should be noted that the p-values computed by any CP will always be valid in the sense of satisfying (6) regardless of the particular algorithm or nonconformity measure definition it uses. The choice of algorithm, nonconformity measure definition and any parameters only affects the tightness of the predictive regions output by the CP, and consequently their usefulness. To demonstrate the influence of an inadequate nonconformity measure definition on the results of a CP, let us consider the case of a trivial definition that always returns the same value $\alpha_i$ for any given example $(x_i, y_i)$. This will make all p-values equal to 1 and will result in the predictive region $\mathbb{R}$ regardless of the required confidence level. Although this region is useless since it does not provide us with any information, it is still valid as it will always contain the true label of the example. Therefore even in the worst case of using some totally wrong nonconformity measure definition or algorithm, the regions produced by the corresponding CP will be useless, but they will never be misleading.

## 3. $k$-Nearest Neighbours Regression TCP

The $k$-Nearest Neighbours algorithms base their predictions on the $k$ training examples that are nearest to the unlabeled example in question according to some distance measure, such as the Euclidean distance. More specifically, for an input vector $x_{l+1}$ the $k$-Nearest Neighbours Regression ($k$-NNR) algorithm finds the $k$ nearest training examples to $x_{l+1}$ and outputs the average (in some cases the median is also used) of their labels as its prediction. A refined form of the method assigns a weight to each one of the $k$ examples depending on their distance from $x_{l+1}$, these weights determine the contribution of each label to the calculation of its prediction; in other words it predicts the weighted average of their labels. It is also worth to mention that the performance of the Nearest Neighbours method can be enhanced by the use of a suitable distance measure or kernel for a specific data set.

Here we will consider the version of the $k$-NNR method which predicts the weighted average of the $k$ nearest examples. In our experiments we used the Euclidean distance, which is the most commonly used distance measure. It will be easy to see that the use of a kernel function or of a different distance measure will not require any changes to our method.

As mentioned in Section 1 in order to create any CP we need to define a nonconformity measure based on the underlying algorithm in question. First let us consider the nonconformity measure

$$\alpha_i = |y_i - \hat{y}_i|, \tag{8}$$

where $\hat{y}_i$ is the prediction of $k$-NNR for $x_i$ based on the examples

$$\{(x_1, y_1), \ldots, (x_{i-1}, y_{i-1}), (x_{i+1}, y_{i+1}), \ldots, (x_{l+1}, \tilde{y})\};$$

recall from Section 2 that $\tilde{y}$ is the assumed label for the new example $x_{l+1}$.

Following Nouretdinov et al. (2001a) and Vovk et al. (2005) we express the nonconformity score $\alpha_i$ of each example $i = 1, \ldots, l+1$ as a piecewise-linear function of $\tilde{y}$

$$\alpha_i = \alpha_i(\tilde{y}) = |a_i + b_i \tilde{y}|.$$

To do this we define $a_i$ and $b_i$ as follows:

- $a_{l+1}$ is minus the weighted average of the labels of the $k$ nearest neighbours of $x_{l+1}$ and $b_{l+1} = 1$;

- if $i \le l$ and $x_{l+1}$ is one of the $k$ nearest neighbours of $x_i$, $a_i$ is $y_i$ minus the labels of the $k-1$ nearest neighbours of $x_i$ from $\{x_1, \ldots, x_{i-1}, x_{i+1}, \ldots, x_l\}$ multiplied by their corresponding weights, and $b_i$ is minus the weight of $x_{l+1}$;

- if $i \le l$ and $x_{l+1}$ is not one of the $k$ nearest neighbours of $x_i$, $a_i$ is $y_i$ minus the weighted average of the labels of the $k$ nearest neighbours of $x_i$, and $b_i = 0$.

As a result the p-value $p(\tilde{y})$ (defined by (5)) corresponding to $\tilde{y}$ can only change at the points where $\alpha_i(\tilde{y}) - \alpha_{l+1}(\tilde{y})$ changes sign for some $i = 1, \ldots, l$. This means that instead of having to calculate the p-value of every possible $\tilde{y}$, we can calculate the set of points $\tilde{y}$ on the real line that have a p-value $p(\tilde{y})$ greater than the given significance level $\delta$, leading to a feasible prediction algorithm.

---

**Algorithm 1**: $k$-NNR TCP

---

**Input**: training set $\{(x_1, y_1), \ldots, (x_l, y_l)\}$, new example $x_{l+1}$, number of nearest
neighbours $k$ and significance level $\delta$.

$P := \{\}$;
**for** $i = 1$ **to** $l + 1$ **do**
    Calculate $a_i$ and $b_i$ for example $z_i = (x_i, y_i)$;
    **if** $b_i < 0$ **then** $a_i := -a_i$ and $b_i := -b_i$;
    **if** $b_i \neq b_{l+1}$ **then** add (10) to $P$;
    **if** $b_i = b_{l+1} \neq 0$ **and** $a_i \neq a_{l+1}$ **then** add (11) to $P$;
**end**
Sort $P$ in ascending order obtaining $y_{(1)}, \ldots, y_{(u)}$;
Add $y_{(0)} := -\infty$ and $y_{(u+1)} := \infty$ to $P$;
$N(j) = 0$, $j = 0, \ldots, u$;
$M(j) = 0$, $j = 1, \ldots, u$;
**for** $i = 1$ **to** $l + 1$ **do**
    **if** $S_i = \{\}$ *(see (9))* **then** Do nothing;
    **else if** $S_i$ *contains only one point,* $S_i = \{y_{(j)}\}$ **then**
        $M(j) := M(j) + 1$;
    **else if** $S_i$ *is an interval* $[y_{(j_1)}, y_{(j_2)}]$, $j_1 < j_2$ **then**
        $M(z) := M(z) + 1, z = j_1, \ldots, j_2$;
        $N(z) := N(z) + 1, z = j_1, \ldots, j_2 - 1$;
    **else if** $S_i$ *is a ray* $(-\infty, y_{(j)}]$ **then**
        $M(z) := M(z) + 1, z = 1, \ldots, j$;
        $N(z) := N(z) + 1, z = 0, \ldots, j - 1$;
    **else if** $S_i$ *is a ray* $[y_{(j)}, \infty)$ **then**
        $M(z) := M(z) + 1, z = j, \ldots, u$;
        $N(z) := N(z) + 1, z = j, \ldots, u$;
    **else if** $S_i$ *is the union* $(-\infty, y_{(j_1)}] \cup [y_{(j_2)}, \infty)$ *of two rays,* $j_1 < j_2$ **then**
        $M(z) := M(z) + 1, z = 1, \ldots, j_1, j_2, \ldots, u$;
        $N(z) := N(z) + 1, z = 0, \ldots, j_1 - 1, j_2, \ldots, u$;
    **else if** $S_i$ *is the real line* $(-\infty, \infty)$ **then**
        $M(z) := M(z) + 1, z = 1, \ldots, u$;
        $N(z) := N(z) + 1, z = 0, \ldots, u$;
**end**
**Output**: the predictive region
    $\left( \cup_{j: \frac{N(j)}{l+1} > \delta} (y_{(j)}, y_{(j+1)}) \right) \cup \{y_{(j)} : \frac{M(j)}{l+1} > \delta\}$.

---

For each $i = 1, \ldots, l+1$, let

$$
\begin{aligned}
S_i &= \{\tilde{y} : \alpha_i(\tilde{y}) \geq \alpha_{l+1}(\tilde{y})\} \\
&= \{\tilde{y} : |a_i + b_i \tilde{y}| \geq |a_{l+1} + b_{l+1} \tilde{y}|\}.
\end{aligned}
\tag{9}
$$

Each set $S_i$ (always closed) will either be an interval, a ray, the union of two rays, the real line, or empty; it can also be a point, which is a special case of an interval. As we are interested in $|a_i + b_i \tilde{y}|$ we can assume that $b_i \geq 0$ for $i = 1, \ldots, l+1$ (if not we multiply both $a_i$ and $b_i$ by $-1$). If $b_i \neq b_{l+1}$, then $\alpha_i(\tilde{y})$ and $\alpha_{l+1}(\tilde{y})$ are equal at two points (which may coincide):

$$
-\frac{a_i - a_{l+1}}{b_i - b_{l+1}} \quad \text{and} \quad -\frac{a_i + a_{l+1}}{b_i + b_{l+1}} ;
\tag{10}
$$

in this case $S_i$ is an interval (maybe a point) or the union of two rays. If $b_i = b_{l+1} \neq 0$, then $\alpha_i(\tilde{y}) = \alpha_{l+1}(\tilde{y})$ at just one point:

$$
-\frac{a_i + a_{l+1}}{2 b_i},
\tag{11}
$$

and $S_i$ is a ray, unless $a_i = a_{l+1}$ in which case $S_i$ is the real line. If $b_i = b_{l+1} = 0$, then $S_i$ is either empty or the real line.

To calculate the p-value $p(\tilde{y})$ for any potential label $\tilde{y}$ of the new example $x_{l+1}$, we count how many $S_i$ include $\tilde{y}$ and divide by $l+1$,

$$
p(\tilde{y}) = \frac{\#\{i = 1, \ldots, l+1 : \tilde{y} \in S_i\}}{l+1}.
\tag{12}
$$

As $\tilde{y}$ increases $p(\tilde{y})$ can only change at the points (10) and (11), so for any significance level $\delta$ we can find the set of $\tilde{y}$ for which $p(\tilde{y}) > \delta$ as the union of finitely many intervals and rays. Algorithm 1 implements a slightly modified version of this idea. It creates a list of the points (10) and (11), sorts it in ascending order obtaining $y_{(1)}, \ldots, y_{(u)}$, adds $y_{(0)} = -\infty$ to the beginning and $y_{(u+1)} = \infty$ to the end of this list, and then computes $N(j)$, the number of $S_i$ which contain the interval $(y_{(j)}, y_{(j+1)})$, for $j = 0, \ldots, u$, and $M(j)$ the number of $S_i$ which contain the point $y_{(j)}$, for $j = 1, \ldots, u$.

## 4. $k$-Nearest Neighbours Regression ICP

As the TCP technique follows a transductive approach, most of its computations are repeated for every test example. The reason for this is that the test example is included in the training set of the underlying algorithm of the TCP in order to calculate the required nonconformity measures. This means that the underlying algorithm is retrained for every test example, which renders TCP quite computationally inefficient for application to large data sets.

Inductive Conformal Predictors (ICP) are based on the same general idea described in Section 2, but follow an inductive approach, which allows them to train their underlying algorithm just once. This is achieved by splitting the training set (of size $l$) into two smaller sets, the *calibration set* with $q < l$ examples and the *proper training set* with $m := l - q$ examples. The proper training set is used for creating the prediction rule $D_{\{z_1, \ldots, z_m\}}$ and only

---

**Algorithm 2**: $k$-NNR ICP

---

**Input**: training set $\{(x_1, y_1), \ldots, (x_l, y_l)\}$, test set $\{x_{l+1}, \ldots, x_{l+r}\}$, number of nearest neighbours $k$, number of calibration examples $q$, and significance level $\delta$.

$m := l - q$;

$P := \{\}$;

**for** $i = 1$ **to** $q$ **do**

    Calculate $\hat{y}_{m+i}$ using $\{(x_1, y_1), \ldots, (x_m, y_m)\}$ as training set;

    Calculate $\alpha_{m+i}$ for the pair $z_{m+i} = (x_{m+i}, y_{m+i})$;

    Add $\alpha_{m+i}$ to $P$;

**end**

Sort $P$ in descending order obtaining $\alpha_{(m+1)}, \ldots, \alpha_{(m+q)}$;

$s := \lfloor \delta(q+1) \rfloor$;

**for** $g = 1$ **to** $r$ **do**

    Calculate $\hat{y}_{l+g}$ using $\{(x_1, y_1), \ldots, (x_m, y_m)\}$ as training set;

    **Output**: the predictive region $(\hat{y}_{l+g} - \alpha_{(m+s)}, \hat{y}_{l+g} + \alpha_{(m+s)})$.

**end**

---

the examples in the calibration set are used for calculating the p-value of each possible label of the new test example. More specifically, the non-conformity score $\alpha_{m+i}$ of each example $z_{m+i}$ in the calibration set $\{z_{m+1}, \ldots, z_{m+q}\}$ is calculated as the degree of disagreement between the prediction

$$\hat{y}_{m+i} = D_{\{z_1, \ldots, z_m\}}(x_{m+i}) \tag{13}$$

and the true label $y_{m+i}$. In the same way, the non-conformity score $\alpha_{l+g}(\tilde{y})$ for the assumed label $\tilde{y}$ of the new test example $x_{l+g}$ is calculated as the degree of disagreement between

$$\hat{y}_{l+g} = D_{\{z_1, \ldots, z_m\}}(x_{l+g}) \tag{14}$$

and $\tilde{y}$. Notice that the nonconformity scores of the examples in the calibration set only need to be computed once. Using these non-conformity scores the p-value of each possible label $\tilde{y}$ of $x_{l+g}$ can be calculated as

$$p(\tilde{y}) = \frac{\#\{i = m+1, \ldots, m+q, l+g : \alpha_i \geq \alpha_{l+g}\}}{q+1}. \tag{15}$$

As with the original CP approach it is impossible to explicitly consider every possible label $\tilde{y} \in \mathbb{R}$ of a new example $x_{l+g}$ and calculate its p-value. However, now both the nonconformity scores of the calibration set examples $\alpha_{m+1}, \ldots, \alpha_{m+q}$ and the $k$-NNR prediction $\hat{y}_{l+g}$ remain fixed for each test example $x_{l+g}$, and the only thing that changes for different values of the assumed label $\tilde{y}$ is the nonconformity score $\alpha_{l+g}$. Therefore $p(\tilde{y})$ changes only at the points where $\alpha_{l+g}(\tilde{y}) = \alpha_i$ for some $i = m+1, \ldots, m+q$. As a result, for a confidence level $1 - \delta$ we only need to find the biggest $\alpha_i$ such that when $\alpha_{l+g}(\tilde{y}) = \alpha_i$ then $p(\tilde{y}) > \delta$, which will give us the maximum and minimum $\tilde{y}$ that have a p-value bigger than $\delta$ and consequently the beginning and end of the corresponding predictive region. More specifically, we sort the nonconformity scores of the calibration examples in descending order obtaining the sequence

$$\alpha_{(m+1)}, \ldots, \alpha_{(m+q)}, \tag{16}$$

and output the predictive region

$$(\hat{y}_{l+g} - \alpha_{(m+s)}, \hat{y}_{l+g} + \alpha_{(m+s)}), \tag{17}$$

where

$$s = \lfloor \delta(q+1) \rfloor. \tag{18}$$

The whole process is detailed in Algorithm 2. Notice that as opposed to Algorithm 1 where all computations have to be repeated for every test example, here only the part inside the second for loop is repeated.

The parameter $q$ given as input to Algorithm 2 determines the number of training examples that will be allocated to the calibration set and the nonconformity scores of which will be used by the ICP to generate its predictive regions. These examples should only take up a small portion of the training set, so that their removal will not dramatically reduce the predictive ability of the underlying algorithm. As we are mainly interested in the confidence levels of 99% and 95%, the calibration sizes we use are of the form $q = 100n - 1$, where $n$ is a positive integer (see (18)).

## 5. Normalized Nonconformity Measures

The main aim of this work was to improve the typical regression nonconformity measure (8) by normalizing it with the expected accuracy of the underlying method. The intuition behind this is that if two examples have the same nonconformity score as defined by (8) and the prediction $\hat{y}$ for one of them was expected to be more accurate than the other, then the former is actually stranger than the latter. This leads to predictive regions that are larger for the examples which are more difficult to predict and smaller for the examples which are easier to predict.

The first measure of expected accuracy we use is based on the distance of the example from its $k$ nearest neighbours. Since the $k$ nearest training examples are the ones actually used to derive the prediction of our underlying method for an example, the nearer these are to the example, the more accurate we expect this prediction to be.

For each example $z_i$, let us denote by $T_i$ the training set used for generating the prediction $\hat{y}_i$. This will be the set

$$T_i = \{z_1, \ldots, z_{i-1}, z_{i+1}, \ldots, z_{l+1}\} \tag{19}$$

in the case of the TCP and the set

$$T_i = \{z_1, \ldots, z_m\} \tag{20}$$

in the case of the ICP. Furthermore, we denote the $k$ nearest neighbours of $x_i$ in $T_i$ as

$$(x_{i_1}, y_{i_1}), \ldots, (x_{i_k}, y_{i_k}). \tag{21}$$

and the sum of the distances between $x_i$ and its $k$ nearest neighbours as

$$d_i^k = \sum_{j=1}^{k} \text{distance}(x_i, x_{i_j}).$$ (22)

We could use $d_i^k$ as a measure of accuracy, in fact it was used successfully in our previous work (Papadopoulos et al., 2008). However, here in order to make this measure more consistent across different data sets we use

$$\lambda_i^k = \frac{d_i^k}{\text{median}(\{d_j^k : z_j \in T_i\})},$$ (23)

which compares the distance of the example from its $k$ nearest neighbours with the median of the distances of all training examples from their $k$ nearest neighbours. Using $\lambda_i^k$ we defined the nonconformity measures:

$$\alpha_i = \left| \frac{y_i - \hat{y}_i}{\gamma + \lambda_i^k} \right|,$$ (24)

and

$$\alpha_i = \left| \frac{y_i - \hat{y}_i}{\exp(\gamma \lambda_i^k)} \right|,$$ (25)

where the parameter $\gamma \geq 0$ controls the sensitivity of each measure to changes of $\lambda_i^k$; in the first case increasing $\gamma$ results in a less sensitive nonconformity measure, while in the second increasing $\gamma$ results in a more sensitive measure. The exponential function in definition (25) was chosen because it has a minimum value of 1, since $\lambda_i^k$ will always be positive, and grows quickly as $\lambda_i^k$ increases. As a result, this measure is more sensitive to changes when $\lambda_i^k$ is big, which indicates that an example is unusually far from the training examples.

The second measure of accuracy we use is based on how different the labels of the example's $k$ nearest neighbours are, which is measured as their standard deviation. The more these labels agree with each other, the more accurate we expect the prediction of the $k$-nearest neighbours algorithm to be. For an example $x_i$, we measure the standard deviation of the labels of its $k$ neighbours as

$$s_i^k = \sqrt{\frac{1}{k} \sum_{j=1}^{k} (y_{i_j} - \overline{y_{i_{1,\ldots,k}}})^2},$$ (26)

where

$$\overline{y_{i_{1,\ldots,k}}} = \frac{1}{k} \sum_{j=1}^{k} y_{i_j}.$$ (27)

Again to make this measure consistent across data sets we divide it with the median standard deviation of the $k$ nearest neighbour labels of all training examples

$$\xi_i^k = \frac{s_i^k}{\text{median}(\{s_j^k : z_j \in T_i\})}.$$ (28)

In the same fashion as (24) and (25) we defined the nonconformity measures:

$$\alpha_i = \left| \frac{y_i - \hat{y}_i}{\gamma + \xi_i^k} \right|, \tag{29}$$

and

$$\alpha_i = \left| \frac{y_i - \hat{y}_i}{\exp(\gamma \xi_i^k)} \right|, \tag{30}$$

where again the parameter $\gamma$ controls the sensitivity of each measure to changes of $\xi_i^k$.

Finally by combining $\lambda_i^k$ and $\xi_i^k$ we defined the nonconformity measures:

$$\alpha_i = \left| \frac{y_i - \hat{y}_i}{\gamma + \lambda_i^k + \xi_i^k} \right|, \tag{31}$$

and

$$\alpha_i = \left| \frac{y_i - \hat{y}_i}{\exp(\gamma \lambda_i^k) + \exp(\rho \xi_i^k)} \right|, \tag{32}$$

where in (31) the parameter $\gamma$ controls the sensitivity of the measure to changes of both $\lambda_i^k$ and $\xi_i^k$, whereas in (32) there are two parameters $\gamma$ and $\rho$, which control the sensitivity to changes of $\lambda_i^k$ and $\xi_i^k$ respectively.

In order to use these nonconformity measures with the $k$-Nearest Neighbours Regression TCP we need to calculate their nonconformity scores as $\alpha_i = |a_i + b_i \tilde{y}|$. We can easily do this for (24) and (25) by computing $a_i$ and $b_i$ as defined in Section 3 and then dividing both by $(\gamma + \lambda_i^k)$ for nonconformity measure (24) and by $\exp(\gamma \lambda_i^k)$ for nonconformity measure (25). Unfortunately however, the same cannot be applied for all other nonconformity measures we defined since $\xi_i^k$ depends on the labels of the $k$ nearest examples, which change for the $k$ nearest neighbours of $x_{l+1}$ as we change $\tilde{y}$. For this reason TCP is limited to using only nonconformity measures (24) and (25).

In the case of the ICP we calculate the nonconformity scores $\alpha_{m+1}, \ldots, \alpha_{m+q}$ of the calibration examples using (24), (25), (29), (30), (31) or (32) and instead of the predictive region (17) we output

$$(\hat{y}_{l+g} - \alpha_{(m+s)}(\gamma + \lambda_i^k), \hat{y}_{l+g} + \alpha_{(m+s)}(\gamma + \lambda_i^k)), \tag{33}$$

for (24),

$$(\hat{y}_{l+g} - \alpha_{(m+s)}\exp(\gamma \lambda_i^k), \hat{y}_{l+g} + \alpha_{(m+s)}\exp(\gamma \lambda_i^k)), \tag{34}$$

for (25),

$$(\hat{y}_{l+g} - \alpha_{(m+s)}(\gamma + \xi_i^k), \hat{y}_{l+g} + \alpha_{(m+s)}(\gamma + \xi_i^k)), \tag{35}$$

for (29),

$$(\hat{y}_{l+g} - \alpha_{(m+s)}\exp(\gamma \xi_i^k), \hat{y}_{l+g} + \alpha_{(m+s)}\exp(\gamma \xi_i^k)), \tag{36}$$

for (30),

$$(\hat{y}_{l+g} - \alpha_{(m+s)}(\gamma + \lambda_i^k + \xi_i^k), \hat{y}_{l+g} + \alpha_{(m+s)}(\gamma + \lambda_i^k + \xi_i^k)), \tag{37}$$

for (31) and

$$(\hat{y}_{l+g} - \alpha_{(m+s)}(\exp(\gamma \lambda_i^k) + \exp(\rho \xi_i^k)), \hat{y}_{l+g} + \alpha_{(m+s)}(\exp(\gamma \lambda_i^k) + \exp(\rho \xi_i^k))), \tag{38}$$

for (32).

## 6. Theoretical Analysis of Nonconformity Measure (29)

In this section we examine $k$-NNR ICP with nonconformity measure (29) under some specific assumptions and show that, under these assumptions, the predictive regions produced are asymptotically optimal; it is important to note that these assumptions are not required for the validity of the resulting predictive regions. We chose not to formalize all the conditions needed for our conclusions, as this would have made our statement far too complicated.

Assume that each label $y_i$ is generated by a normal distribution $\mathcal{N}(\mu_{x_i}, \sigma_{x_i}^2)$, where $\mu_x$ and $\sigma_x$ are smooth functions of $x$, that each $x_i$ is generated by a probability distribution that is concentrated on a compact set and whose density is always greater than some constant $\epsilon > 0$, and that $k \gg 1$, $m \gg k$ and $q \gg k$. In this case nonconformity measure (29) with $\gamma = 0$ will be

$$\alpha_i = \left| \frac{y_i - \hat{y}_i}{s_i^k} \right| \approx \left| \frac{y_i - \mu_{x_i}}{\sigma_{x_i}} \right|, \tag{39}$$

where the division of $s_i^k$ by median($\{s_j^k : z_j \in T_i\}$) is ignored since the latter does not change within the same data set. The values

$$\frac{y_{m+1} - \mu_{x_{m+1}}}{\sigma_{x_{m+1}}}, \ldots, \frac{y_{m+q} - \mu_{x_{m+q}}}{\sigma_{x_{m+q}}}$$

will follow an approximately standard normal distribution, and for a new example $x_{l+g}$ with probability close to $1 - \delta$ we have

$$\frac{y_{l+g} - \mu_{x_{l+g}}}{\sigma_{x_{l+g}}} \in \left[ -\alpha_{(m+\lfloor \delta q \rfloor)}, \alpha_{(m+\lfloor \delta q \rfloor)} \right],$$

where $\alpha_{(m+1)}, \ldots, \alpha_{(m+q)}$ are the nonconformity scores $\alpha_{m+1}, \ldots, \alpha_{m+q}$ sorted in descending order. As a result we obtain the region

$$y_{l+g} \in \left[ \mu_{x_{l+g}} - \alpha_{(m+\lfloor \delta q \rfloor)} \sigma_{x_{l+g}}, \mu_{x_{l+g}} + \alpha_{(m+\lfloor \delta q \rfloor)} \sigma_{x_{l+g}} \right],$$

which, on one hand, is close to the standard (and optimal in various senses) prediction interval for the normal model and, on the other hand, is almost identical to the region (35) of $k$-NNR ICP (recall that we set $\gamma = 0$ and $\xi_i^k = s_i^k$).

## 7. Experimental Results

Our methods were tested on six benchmark data sets from the UCI (Frank & Asuncion, 2010) and DELVE (Rasmussen et al., 1996) repositories:

- *Boston Housing*, which lists the median house prices for 506 different areas of Boston MA in $1000s. Each area is described by 13 attributes such as pollution and crime rate.

- *Abalone*, which concerns the prediction of the age of abalone from physical measurements. The data set consists of 4177 examples described by 8 attributes such as diameter, height and shell weight.

- *Computer Activity*, which is a collection of a computer systems activity measures from a Sun SPARCstation 20/712 with 128 Mbytes of memory running in a multi-user university department. It consists of 8192 examples of 12 measured values, such as the number of system buffer reads per second and the number of system call writes per second, at random points in time. The task is to predict the portion of time that the cpus run in user mode, ranging from 0 to 100. We used the *small* variant of the data set which contains only 12 of the 21 attributes.

- *Kin*, which was generated from a realistic simulation of the forward dynamics of an 8 link all-revolute robot arm. The task is to predict the distance of the end-effector from a target. The data set consists of 8192 examples described by attributes like joint positions and twist angles. We used the *8nm* variant of the data set which contains 8 of the 32 attributes, and is highly non-linear with moderate noise.

- *Bank*, which was generated from simplistic simulator of the queues in a series of banks. The task is to predict the rate of rejections, i.e. the fraction of customers that are turned away from the bank because all the open tellers have full queues. The data set consists of 8192 examples described by 8 attributes like area population size and maximum possible length of queues. The *8nm* variant of the data set was used with the same characteristics given in the description of the *Kin* data set.

- *Pumadyn*, which was generated from a realistic simulation of the dynamics of a Unimation Puma 560 robot arm. It consists of 8192 examples and the task is to predict the angular acceleration of one of the robot arm's links. Each example is described by 8 attributes, which include angular positions, velocities and torques of the robot arm. The *8nm* variant of the data set was used with the same characteristics given in the description of the *Kin* data set.

Before conducting our experiments the attributes of all data sets were normalized to a minimum value of 0 and a maximum value of 1. Our experiments consisted of 10 random runs of a fold cross-validation process. Based on their sizes the Boston Housing and Abalone data sets were split into 10 and 4 folds respectively, while the other four were splint into 2 folds. For determining the number $k$ of nearest neighbours that was used for each data set, one third of the training set of its first fold was held-out as a validation set and the base algorithm was tested on that set with different $k$, using the other two thirds for training. The number of neighbours $k$ that gave the smallest mean absolute error was selected. Note that, as explained in Section 2, the choice of $k$ and any other parameter does not affect the validity of the results produced by the corresponding CP, it only affects their efficiency. The calibration set sizes were set to $q = 100n - 1$ (see Section 4), where $n$ was chosen so that $q$ was approximately 1/10th of each data set's training size; in the case of the Boston Housing data set the smallest value $n = 1$ was used. Table 1 gives the number of folds, number of nearest neighbours $k$, and calibration set size $q$ used in our experiments for each data set, together with the number of examples and attributes it consists of and the width of its range of labels.

The parameters $\gamma$ and $\rho$ of our nonconformity measures were set in all cases to 0.5, which seems to give very good results with all data sets and measures. It is worth to note however, that somewhat tighter predictive regions can be obtained by adjusting the corresponding parameter(s) of each measure for each data set. We chose to fix these parameters to 0.5 here, so as to show that the remarkable improvement in the predictive region widths resulting from the use of the new nonconformity measures does not depend on fine tuning these parameters.

| | Boston Housing | Abalone | Computer Activity | Kin | Bank | Pumadyn |
|---|---|---|---|---|---|---|
| Examples | 506 | 4177 | 8192 | 8192 | 8192 | 8192 |
| Attributes | 13 | 8 | 12 | 8 | 8 | 8 |
| Label range | 45 | 28 | 99 | 1.42 | 0.48 | 21.17 |
| Folds | 10 | 4 | 2 | 2 | 2 | 2 |
| $k$ | 4 | 16 | 8 | 7 | 4 | 6 |
| Calibration size | 99 | 299 | 399 | 399 | 399 | 399 |

Table 1: Main characteristics and experimental setup for each data set.

Since our methods output predictive regions instead of point predictions, the main aim of our experiments was to check the tightness of these regions. The first two parts of Tables 2-7 report the median and interdecile mean widths of the regions produced for every data set by each nonconformity measure of the $k$-NNR TCP and ICP for the 99%, 95% and 90% confidence levels. We chose to report the median and interdecile mean values instead of the mean so as to avoid the strong impact of a few extremely large or extremely small regions.

In the third and last parts of Tables 2-7 we check the reliability of the obtained predictive regions for each data set. This is done by reporting the percentage of examples for which the true label is not inside the region output by the corresponding method. In effect this checks empirically the validity of the predictive regions. The percentages reported here are very close to the required significance levels and do not change by much for the different nonconformity measures.

Figures 1-6 complement the information detailed in Tables 2-7 by displaying boxplots which show the median, upper and lower quartiles, and upper and lower deciles of the predictive region widths produced for each data set. Each chart is divided into three parts, separating the three confidence levels we consider, and each part contains 10 boxplots of which the first three are for the $k$-NNR TCP with the nonconformity measures (8), (24) and (25), and the remaining seven are for the $k$-NNR ICP with all nonconformity measures.

A transformation of the width values reported in Tables 2-7 to the percentage of the range of possible labels they represent shows that in general the predictive regions produced by all our methods are relatively tight. The median width percentages of all nonconformity measures and all data sets are between 17% and 86% for the 99% confidence level and between 11% and 47% for the 95% confidence level. If we now consider the best performing nonconformity measure for each data set, the worst median width percentage for the 99% confidence level is 61% and for the 95% confidence level it is 43% (both for the pumadyn data set).

By comparing the predictive region tightness of the different nonconformity measures for each method both in Tables 2-7 and in Figures 1-6, one can see the relatively big improvement that our new nonconformity measures achieve as compared to the standard

| Method/ Measure | | Median Width | | | Interdecile Mean Width | | | Percentage outside predictive regions | | |
|---|---|---|---|---|---|---|---|---|---|---|
| | | 90% | 95% | 99% | 90% | 95% | 99% | 90% | 95% | 99% |
| TCP | (8) | 12.143 | 17.842 | 33.205 | 12.054 | 17.870 | 33.565 | 10.24 | 5.06 | 0.97 |
| | (24) | 11.172 | 14.862 | 24.453 | 11.483 | 15.289 | 25.113 | 9.94 | 4.80 | 0.89 |
| | (25) | 10.897 | 14.468 | 24.585 | 11.258 | 14.956 | 25.368 | 9.92 | 4.92 | 0.91 |
| ICP | (8) | 13.710 | 19.442 | 38.808 | 13.693 | 19.417 | 41.581 | 9.47 | 4.88 | 0.79 |
| | (24) | 11.623 | 16.480 | 30.427 | 11.985 | 16.991 | 33.459 | 10.59 | 4.92 | 0.71 |
| | (25) | 11.531 | 16.702 | 30.912 | 11.916 | 17.116 | 34.477 | 10.08 | 4.72 | 0.69 |
| | (29) | 11.149 | 15.233 | 36.661 | 12.165 | 16.645 | 41.310 | 9.55 | 4.82 | 0.59 |
| | (30) | 10.211 | 14.228 | 34.679 | 11.347 | 15.820 | 39.211 | 9.68 | 4.60 | 0.65 |
| | (31) | 10.712 | 14.723 | 28.859 | 11.343 | 15.612 | 31.120 | 9.88 | 4.76 | 0.61 |
| | (32) | 10.227 | 13.897 | 29.068 | 10.876 | 14.832 | 31.810 | 9.31 | 4.88 | 0.57 |

Table 2: The tightness and reliability results of our methods on the Boston Housing data set.

| Method/ Measure | | Median Width | | | Interdecile Mean Width | | | Percentage outside predictive regions | | |
|---|---|---|---|---|---|---|---|---|---|---|
| | | 90% | 95% | 99% | 90% | 95% | 99% | 90% | 95% | 99% |
| TCP | (8) | 6.685 | 9.267 | 16.109 | 6.712 | 9.259 | 16.124 | 9.94 | 4.94 | 0.95 |
| | (24) | 6.213 | 8.487 | 14.211 | 6.376 | 8.708 | 14.581 | 10.03 | 4.98 | 0.98 |
| | (25) | 6.034 | 8.278 | 13.949 | 6.230 | 8.545 | 14.393 | 10.03 | 4.95 | 0.99 |
| ICP | (8) | 6.705 | 9.486 | 16.628 | 6.671 | 9.388 | 16.580 | 10.32 | 5.09 | 1.01 |
| | (24) | 6.200 | 8.305 | 14.012 | 6.359 | 8.513 | 14.520 | 10.57 | 5.54 | 1.20 |
| | (25) | 6.057 | 8.205 | 13.922 | 6.229 | 8.434 | 14.457 | 10.50 | 5.46 | 1.19 |
| | (29) | 5.837 | 7.987 | 14.394 | 6.004 | 8.229 | 14.895 | 10.47 | 5.05 | 1.02 |
| | (30) | 5.731 | 7.926 | 14.631 | 5.931 | 8.200 | 15.173 | 10.37 | 5.00 | 0.93 |
| | (31) | 5.936 | 7.999 | 13.999 | 6.070 | 8.174 | 14.406 | 10.57 | 5.35 | 1.10 |
| | (32) | 5.838 | 7.962 | 14.028 | 5.994 | 8.178 | 14.506 | 10.54 | 5.23 | 1.09 |

Table 3: The tightness and reliability results of our methods on the Abalone data set.

| Method/ Measure | | Median Width | | | Interdecile Mean Width | | | Percentage outside predictive regions | | |
|---|---|---|---|---|---|---|---|---|---|---|
| | | 90% | 95% | 99% | 90% | 95% | 99% | 90% | 95% | 99% |
| TCP | (8) | 10.005 | 13.161 | 21.732 | 10.009 | 13.113 | 21.679 | 9.98 | 4.99 | 0.97 |
| | (24) | 8.788 | 11.427 | 18.433 | 9.238 | 12.017 | 19.372 | 9.95 | 4.92 | 0.95 |
| | (25) | 8.370 | 10.856 | 17.084 | 8.924 | 11.572 | 18.207 | 9.92 | 4.91 | 0.97 |
| ICP | (8) | 10.149 | 13.588 | 22.705 | 10.245 | 13.467 | 22.577 | 9.71 | 4.79 | 0.95 |
| | (24) | 9.024 | 11.725 | 18.948 | 9.483 | 12.333 | 19.918 | 9.51 | 4.72 | 0.90 |
| | (25) | 8.646 | 11.340 | 17.817 | 9.206 | 12.067 | 18.953 | 9.40 | 4.51 | 0.92 |
| | (29) | 8.837 | 11.877 | 19.595 | 9.031 | 12.145 | 20.114 | 9.75 | 4.59 | 0.91 |
| | (30) | 8.702 | 11.618 | 18.859 | 9.013 | 12.031 | 19.522 | 9.49 | 4.58 | 0.95 |
| | (31) | 8.653 | 11.301 | 18.179 | 9.020 | 11.789 | 18.983 | 9.70 | 4.55 | 0.89 |
| | (32) | 8.517 | 11.114 | 17.468 | 8.914 | 11.627 | 18.264 | 9.46 | 4.51 | 0.96 |

Table 4: The tightness and reliability results of our methods on the Computer Activity data set.

| Method/ Measure | | Median Width | | | Interdecile Mean Width | | | Percentage outside predictive regions | | |
|---|---|---|---|---|---|---|---|---|---|---|
| | | 90% | 95% | 99% | 90% | 95% | 99% | 90% | 95% | 99% |
| TCP | (8) | 0.402 | 0.491 | 0.675 | 0.402 | 0.491 | 0.675 | 10.13 | 4.99 | 0.96 |
| | (24) | 0.395 | 0.480 | 0.649 | 0.396 | 0.481 | 0.651 | 10.18 | 5.01 | 0.95 |
| | (25) | 0.396 | 0.481 | 0.653 | 0.397 | 0.482 | 0.655 | 10.16 | 5.00 | 0.96 |
| ICP | (8) | 0.413 | 0.508 | 0.705 | 0.414 | 0.515 | 0.702 | 9.86 | 4.52 | 0.81 |
| | (24) | 0.408 | 0.498 | 0.680 | 0.409 | 0.499 | 0.682 | 9.73 | 4.61 | 0.77 |
| | (25) | 0.408 | 0.501 | 0.681 | 0.408 | 0.502 | 0.682 | 9.84 | 4.59 | 0.80 |
| | (29) | 0.412 | 0.497 | 0.730 | 0.418 | 0.504 | 0.741 | 9.74 | 5.21 | 0.91 |
| | (30) | 0.401 | 0.482 | 0.695 | 0.408 | 0.491 | 0.707 | 9.74 | 5.08 | 0.84 |
| | (31) | 0.403 | 0.486 | 0.677 | 0.406 | 0.489 | 0.682 | 9.81 | 4.83 | 0.87 |
| | (32) | 0.399 | 0.487 | 0.670 | 0.402 | 0.490 | 0.676 | 10.05 | 4.75 | 0.85 |

Table 5: The tightness and reliability results of our methods on the Kin data set.

| Method/ Measure | | Median Width | | | Interdecile Mean Width | | | Percentage outside predictive regions | | |
|---|---|---|---|---|---|---|---|---|---|---|
| | | 90% | 95% | 99% | 90% | 95% | 99% | 90% | 95% | 99% |
| TCP | (8) | 0.135 | 0.201 | 0.342 | 0.135 | 0.201 | 0.341 | 10.05 | 4.99 | 0.95 |
| | (24) | 0.121 | 0.170 | 0.265 | 0.122 | 0.171 | 0.267 | 10.02 | 4.94 | 0.90 |
| | (25) | 0.123 | 0.173 | 0.270 | 0.124 | 0.175 | 0.273 | 10.03 | 4.95 | 0.91 |
| ICP | (8) | 0.142 | 0.209 | 0.361 | 0.139 | 0.209 | 0.362 | 9.95 | 4.71 | 0.83 |
| | (24) | 0.122 | 0.178 | 0.274 | 0.123 | 0.179 | 0.277 | 10.16 | 4.71 | 0.87 |
| | (25) | 0.125 | 0.182 | 0.282 | 0.126 | 0.183 | 0.284 | 10.08 | 4.72 | 0.87 |
| | (29) | 0.096 | 0.151 | 0.306 | 0.110 | 0.174 | 0.352 | 9.97 | 4.51 | 0.80 |
| | (30) | 0.084 | 0.137 | 0.272 | 0.104 | 0.169 | 0.334 | 10.21 | 4.61 | 0.79 |
| | (31) | 0.096 | 0.146 | 0.247 | 0.105 | 0.160 | 0.270 | 10.14 | 4.46 | 0.85 |
| | (32) | 0.092 | 0.143 | 0.249 | 0.102 | 0.159 | 0.274 | 10.31 | 4.58 | 0.88 |

Table 6: The tightness and reliability results of our methods on the Bank data set.

| Method/ Measure | | Median Width | | | Interdecile Mean Width | | | Percentage outside predictive regions | | |
|---|---|---|---|---|---|---|---|---|---|---|
| | | 90% | 95% | 99% | 90% | 95% | 99% | 90% | 95% | 99% |
| TCP | (8) | 7.909 | 9.694 | 13.908 | 7.927 | 9.726 | 13.827 | 9.92 | 4.87 | 0.99 |
| | (24) | 7.761 | 9.417 | 13.153 | 7.781 | 9.439 | 13.183 | 9.88 | 4.98 | 1.04 |
| | (25) | 7.774 | 9.463 | 13.239 | 7.793 | 9.486 | 13.268 | 9.92 | 4.94 | 1.02 |
| ICP | (8) | 8.008 | 10.005 | 13.924 | 8.097 | 9.993 | 14.007 | 9.72 | 4.75 | 1.06 |
| | (24) | 7.869 | 9.627 | 13.193 | 7.894 | 9.655 | 13.233 | 9.86 | 4.81 | 1.09 |
| | (25) | 7.892 | 9.713 | 13.335 | 7.919 | 9.731 | 13.377 | 9.89 | 4.75 | 1.10 |
| | (29) | 7.745 | 9.563 | 14.342 | 7.842 | 9.698 | 14.588 | 9.96 | 4.87 | 1.04 |
| | (30) | 7.460 | 9.199 | 13.398 | 7.634 | 9.410 | 13.723 | 9.97 | 4.82 | 1.07 |
| | (31) | 7.549 | 9.185 | 13.040 | 7.610 | 9.263 | 13.180 | 10.02 | 4.89 | 1.04 |
| | (32) | 7.579 | 9.237 | 12.808 | 7.660 | 9.336 | 12.955 | 9.90 | 4.83 | 1.15 |

Table 7: The tightness and reliability results of our methods on the Pumadyn data set.

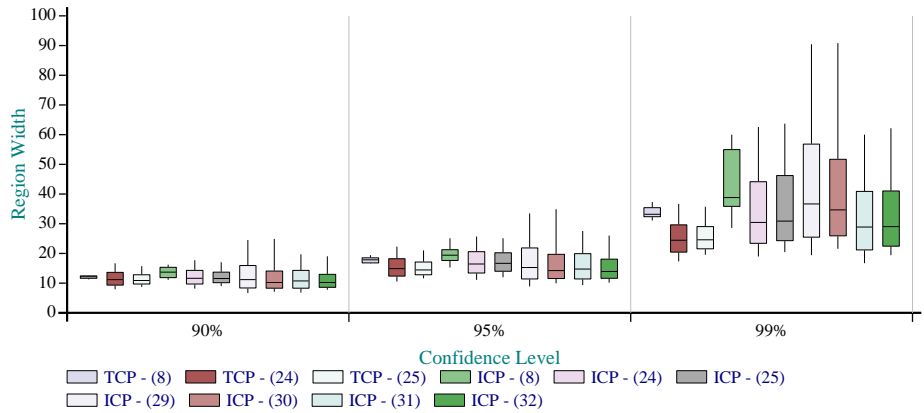

Figure 1: Predictive region width distribution for the Boston housing data set.

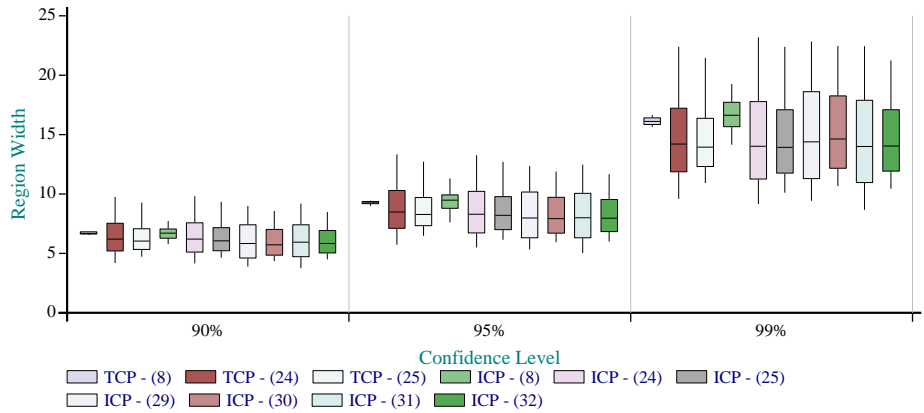

Figure 2: Predictive region width distribution for the Abalone data set.

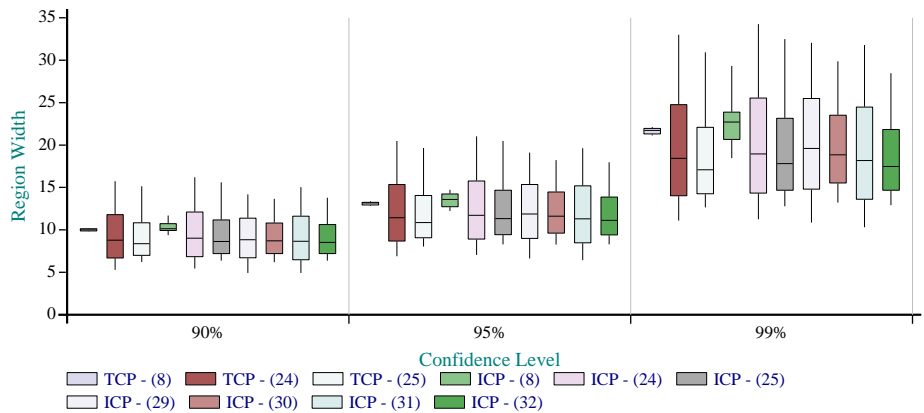

Figure 3: Predictive region width distribution for the Computer Activity data set.

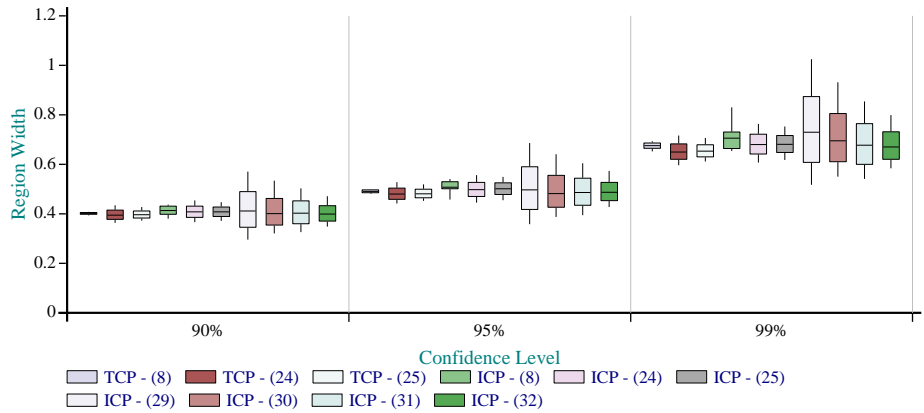

Figure 4: Predictive region width distribution for the Kin data set.

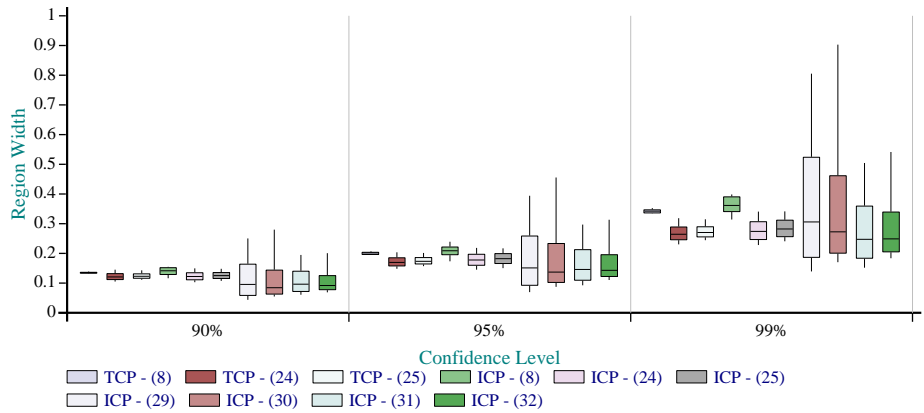

Figure 5: Predictive region width distribution for the Bank data set.

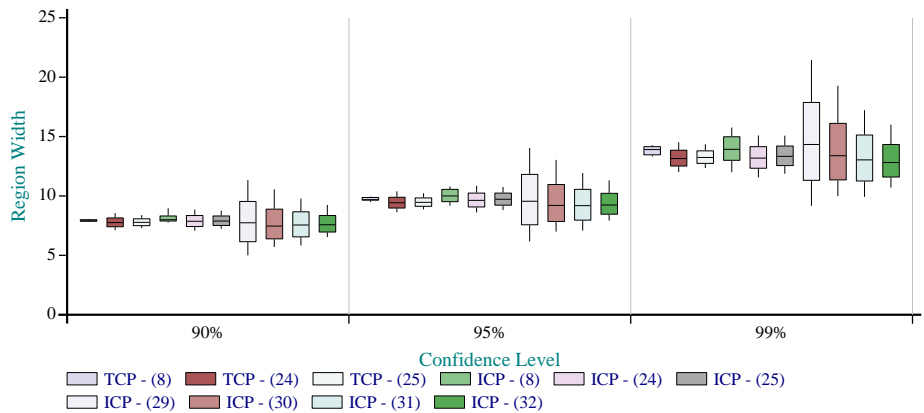

Figure 6: Predictive region width distribution for the Pumadyn data set.

| Method | B. Housing | Abalone | All Other |
|--------|-----------|---------|-----------|
| TCP with all measures | 20 sec | 51 - 52 min | 2.5 - 2.7 hrs |
| ICP with (8) | 0.6 sec | 8 sec | 17 - 18 sec |
| ICP with new measures | 1.4 sec | 29 sec | 37 - 39 sec |

Table 8: The processing times of $k$-NNR TCP and ICP.

regression measure (8). In almost all cases the new measures give smaller median and interdecile mean widths, while in the majority of cases the difference is quite significant. The degree of improvement is more evident in Figures 1-6 where we see that in many cases the median widths of the predictive regions obtained with the new measures are even below the smallest widths obtained with measure (8).

A comparison between the nonconformity measures that are based on the distances of the $k$ nearest neighbours, (24) and (25), and those that are based on the standard deviation of their labels, (29) and (30), reveals that the regions produced by the latter cover a much bigger range of widths. Furthermore, the widths of (24) and (25) seem to be in most cases tighter on average than those of (29) and (30) for the highest confidence level 99%, whereas the opposite is true for the 90% and 95% confidence levels. It is also worth mentioning that measures (29) and (30) are the only ones that produced predictive regions with bigger median and interdecile mean widths than those of measure (8) in some cases.

The last two measures (31) and (32), which combine the others, seem to give the tightest predictive region widths of ICP overall. In 11 out of 18 cases one of the two has the smallest median predictive region width and in all but 1 cases one of the two has the smallest interdecile mean width. Their superiority is also evident in the figures and especially in Figures 1, 5 and 6.

One other important comparison is between the widths of the regions produced by the TCP and ICP approaches. For the standard regression nonconformity measure (8) the regions of the TCP are in all cases tighter than those of the ICP. This is due to the much richer set of examples that TCP uses for calculating its predictive regions; TCP uses the whole training set as opposed to just the calibration examples used by ICP. The difference in predictive region tightness is much bigger in the results of the Boston housing data set, since the calibration set in this case consisted of only 99 examples. It is also worth to note that the widths of the regions produced by ICP with (8) vary much more than the corresponding widths of the TCP, this is natural since the composition of the calibration set changes to a much bigger degree from one run to another than the composition of the whole training set. If we now compare the region widths of the two approaches with nonconformity measures (24) and (25) we see that, with the exception of the Boston housing set, they are very similar in terms of distribution. This shows that the new measures are not so dependant on the composition of the examples used for producing the predictive regions. Furthermore, if we compare the region widths of the TCP with nonconformity measures (24) and (25) with those of the ICP with nonconformity measures (31) and (32), we see that in many cases the regions of the ICP are somewhat tighter, despite the much smaller set of examples it uses to compute them. This is due to the superiority of the measures (31) and (32).

Finally, in Table 8 we report the processing times of the two approaches. We use two rows for reporting the times of the ICP, one for measure (8) and one for all new measures, since the new measures require some extra computations; i.e. finding the distances of all training examples from their $k$ nearest neighbours and/or the standard deviation of their $k$ nearest neighbour labels. We also group the times of the Computer Activity, Kin, Bank and Pumadyn data sets, which were almost identical as they consist of the same number of examples, into one column. This table demonstrates the huge computational efficiency improvement of ICP over TCP. It also shows that there is some computational overhead involved in using the new nonconformity measures with ICP, but it is not so important, especially baring in mind the degree of improvement they bring in terms of predictive region tightness.

## 8. Comparison with Gaussian Process Regression

In this section we compare the predictive regions produced by our methods with those produced by Gaussian Processes (GPs, Rasmussen & Williams, 2006), which is one of the most popular Bayesian machine learning approaches. We first compare Gaussian Process Regression (GPR) and the $k$-NNR CPs on artificially generated data that satisfy the GP prior and then check the results of GPR on three of the data sets described in Section 7, namely Boston Housing, Abalone and Computer Activity. Our implementation of GPR was based on the Matlab code that accompanies the work of Rasmussen and Williams.

For our first set of experiments we generated 100 artificial data sets consisting of 1000 training and 1000 test examples with inputs drawn from a uniform distribution over $[-10, 10]^5$. The labels of each data set were generated from a Gaussian Process with a covariance function defined as the sum of a squared exponential (SE) covariance and independent noise and hyperparameters $(l, \sigma_f, \sigma_n) = (1, 1, 0.1)$; i.e. a unit length scale, a unit signal magnitude and a noise standard deviation of 0.1. We then applied GPR on these data sets using exactly the same covariance function and hyperparameters and compared the results with those of the $k$-NNR CPs (which do not take into account any information about how the data were generated). Table 9 reports the results over all 100 data sets obtained by GPR and our methods in the same manner as Tables 2-7. In this case, since the data meets the GPR prior, the percentage of labels outside the predictive regions produced by GPR are more or less equal to the required significance level as we would expect. The same is true for the regions produced by all our methods. Also, although the predictive regions produced by GPR are tighter than those of produced by our methods, the difference between the two is very small.

Our experiments on the three benchmark data sets were performed following exactly the same setting with our experiments on the two CPs, including the use of the same seed for each fold-cross validation run. In terms of data preprocessing, we normalised the attributes of each data set setting the mean of each attribute to 0 and its standard deviation to 1, while we also centred the labels of each data set so as to have a zero mean; these preprocessing steps are advisable for GPR. We tried out a SE covariance and a Matern covariance with smoothness set to $v = 3/2$ and $v = 5/2$. In the case of the SE covariance, we used the automatic relevance determination version of the function, which allows for a separate length-scale for each attribute determined by a corresponding hyperparameter. In all cases the actual covariance function was defined as the sum of the corresponding covariance and independent noise. All hyperparameters were adapted by maximizing marginal likelihood on each training set as suggested by Rasmussen and Williams (2006); the adaptation of hyperparameters using leave-one-out cross-validation produces more or less the same results.

| Method/ Measure | | Median Width | | | Interdecile Mean Width | | | Percentage outside predictive regions | | |
|---|---|---|---|---|---|---|---|---|---|---|
| | | 90% | 95% | 99% | 90% | 95% | 99% | 90% | 95% | 99% |
| GPR | | 3.306 | 3.940 | 5.178 | 3.306 | 3.939 | 5.177 | 10.01 | 4.95 | 0.99 |
| TCP | (8) | 3.332 | 3.999 | 5.277 | 3.332 | 3.989 | 5.274 | 9.93 | 4.81 | 0.95 |
| | (24) | 3.322 | 3.982 | 5.306 | 3.339 | 4.001 | 5.332 | 9.90 | 4.82 | 0.98 |
| | (25) | 3.322 | 3.975 | 5.290 | 3.335 | 3.990 | 5.309 | 9.88 | 4.82 | 0.95 |
| ICP | (8) | 3.344 | 3.978 | 5.264 | 3.337 | 4.001 | 5.283 | 9.90 | 4.80 | 0.95 |
| | (24) | 3.346 | 4.016 | 5.329 | 3.358 | 4.029 | 5.349 | 9.83 | 4.76 | 1.01 |
| | (25) | 3.337 | 3.994 | 5.302 | 3.348 | 4.005 | 5.320 | 9.86 | 4.78 | 0.98 |
| | (29) | 3.333 | 3.989 | 5.292 | 3.339 | 3.994 | 5.305 | 9.97 | 4.88 | 0.96 |
| | (30) | 3.331 | 3.993 | 5.280 | 3.338 | 3.997 | 5.295 | 9.90 | 4.84 | 0.95 |
| | (31) | 3.340 | 3.995 | 5.308 | 3.348 | 4.002 | 5.324 | 9.85 | 4.79 | 0.97 |
| | (32) | 3.332 | 3.990 | 5.279 | 3.339 | 3.994 | 5.294 | 9.88 | 4.81 | 0.97 |

Table 9: Comparison of our methods with Gaussian Process Regression on artificial data generated by a Gaussian Process.

| Covariance Function | Median Width | | | Interdecile Mean Width | | | Percentage outside predictive regions | | |
|---|---|---|---|---|---|---|---|---|---|
| | 90% | 95% | 99% | 90% | 95% | 99% | 90% | 95% | 99% |
| SE | 7.168 | 8.540 | 11.225 | 7.419 | 8.840 | 11.618 | 10.28 | 6.70 | 2.98 |
| Matern $v = 3/2$ | 8.369 | 9.972 | 13.106 | 8.636 | 10.290 | 13.524 | 8.02 | 5.04 | 2.65 |
| Matern $v = 5/2$ | 8.476 | 10.100 | 13.274 | 8.748 | 10.423 | 13.699 | 7.71 | 4.88 | 2.61 |

Table 10: The tightness and reliability results of Gaussian Process Regression on the Boston Housing data set.

| Covariance Function | Median Width | | | Interdecile Mean Width | | | Percentage outside predictive regions | | |
|---|---|---|---|---|---|---|---|---|---|
| | 90% | 95% | 99% | 90% | 95% | 99% | 90% | 95% | 99% |
| SE | 6.705 | 7.988 | 10.499 | 6.711 | 7.996 | 10.509 | 9.43 | 6.47 | 2.92 |
| Matern $v = 3/2$ | 6.740 | 8.031 | 10.555 | 7.046 | 8.396 | 11.034 | 9.02 | 6.11 | 2.77 |
| Matern $v = 5/2$ | 6.750 | 8.042 | 10.570 | 6.755 | 8.048 | 10.577 | 9.15 | 6.23 | 2.82 |

Table 11: The tightness and reliability results of Gaussian Process Regression on the Abalone data set.

| Covariance Function | Median Width | | | Interdecile Mean Width | | | Percentage outside predictive regions | | |
|---|---|---|---|---|---|---|---|---|---|
| | 90% | 95% | 99% | 90% | 95% | 99% | 90% | 95% | 99% |
| SE | 8.115 | 9.669 | 12.708 | 8.198 | 9.768 | 12.838 | 10.15 | 6.47 | 2.38 |
| Matern $v = 3/2$ | 8.120 | 9.675 | 12.715 | 8.471 | 10.093 | 13.265 | 9.63 | 5.91 | 2.09 |
| Matern $v = 5/2$ | 8.340 | 9.937 | 13.060 | 8.617 | 10.267 | 13.494 | 9.63 | 5.81 | 2.09 |

Table 12: The tightness and reliability results of Gaussian Process Regression on the Computer Activity data set.

Tables 10-12 report the results obtained by GRP on the three data sets with each covariance function. By comparing the values reported in the first two parts of this tables with those in Tables 2-4 one can see that the regions produced by GPR are tighter in almost all cases. However, the percentage of predictive regions that do not include the true label of the example is much higher than the required for the 95% and 99% confidence levels. This shows that the predictive regions produced by GPR are not valid and therefore they are misleading if the correct prior is not known. On the contrary, as demonstrated in Section 2, CPs produce valid predictive regions even if the parameters or underlying algorithm used are totally wrong.

## 9. Conclusions

We presented the Transductive and Inductive Conformal Predictors based on the $k$-Nearest Neighbours Regression algorithm. In addition to the typical regression nonconformity measure, we developed six novel definitions which take into account the expected accuracy of the $k$-NNR algorithm on the example in question. Our definitions assess the expected accuracy of $k$-NNR on the example based on its distances from its $k$ nearest examples (24) and (25), on the standard deviation of their labels (29) and (30), or on a combination of the two (31) and (32).

The experimental results obtained by applying our methods to various data sets show that in all cases they produce reliable predictive intervals that are tight enough to be useful in practice. Additionally, they illustrate the great extent to which our new nonconformity measure definitions improve the performance of both the transductive and inductive method in terms of predictive region tightness. In the case of the ICP, with which all new measures were evaluated, definitions (31) and (32) appear to be superior to all other measures, giving the overall tightest predictive regions. Moreover, a comparison between the TCP and ICP methods suggests that, when dealing with relatively large data sets the use of nonconformity measures (31) and (32) makes ICP perform equally well with TCP in terms of predictive region tightness, whereas it has a vast advantage when it comes to computational efficiency. Finally, a comparison with Gaussian Process Regression (GPR) demonstrated that our methods produce almost as tight predictive regions as those of GPR when the correct prior is known, while GPR may produce misleading regions on real world data on which the required prior knowledge is not available.

The main future direction of this work is the development of normalized nonconformity measures like the ones presented in this paper based on other popular regression techniques, such as Ridge Regression and Support Vector Regression. Although in the case of Ridge Regression one such measure was already defined for ICP (Papadopoulos et al., 2002a), it unfortunately cannot be used with the TCP approach; thus there is potentially a considerable performance gain to be achieved from a definition of this kind for TCP. Moreover, an equally important future aim is the application of our methods to medical or other problems where the provision of predictive regions is desirable, and the evaluation of their results by experts in the corresponding field.

## Acknowledgments

We would like to thank Savvas Pericleous and Haris Haralambous for useful discussions. We would also like to thank the anonymous reviewers for their insightful and constructive comments. This work was supported in part by the Cyprus Research Promotion Foundation through research contract PLHRO/0506/22 ("Development of New Conformal Prediction Methods with Applications in Medical Diagnosis").

## References

Bellotti, T., Luo, Z., Gammerman, A., Delft, F. W. V., & Saha, V. (2005). Qualified predictions for microarray and proteomics pattern diagnostics with confidence machines. *International Journal of Neural Systems*, *15*(4), 247–258.

Cristianini, N., & Shawe-Taylor, J. (2000). *An Introduction to Support Vector Machines and Other Kernel-based Methods*. Cambridge University Press, Cambridge.

Dashevskiy, M., & Luo, Z. (2008). Network traffic demand prediction with confidence. In *Proceedings of the IEEE Global Telecommunications Conference 2008 (GLOBECOM 2008)*, pp. 1453–1457. IEEE.

Frank, A., & Asuncion, A. (2010). UCI machine learning repository. URL http://archive.ics.uci.edu/ml.

Gammerman, A., Vapnik, V., & Vovk, V. (1998). Learning by transduction. In *Proceedings of the Fourteenth Conference on Uncertainty in Artificial Intelligence*, pp. 148–156, San Francisco, CA. Morgan Kaufmann.

Gammerman, A., Vovk, V., Burford, B., Nouretdinov, I., Luo, Z., Chervonenkis, A., Waterfield, M., Cramer, R., Tempst, P., Villanueva, J., Kabir, M., Camuzeaux, S., Timms, J., Menon, U., & Jacobs, I. (2009). Serum proteomic abnormality predating screen detection of ovarian cancer. *The Computer Journal*, *52*(3), 326–333.

Gammerman, A., & Vovk, V. (2007). Hedging predictions in machine learning: The second *computer journal* lecture. *The Computer Journal*, *50*(2), 151–163.

Holst, H., Ohlsson, M., Peterson, C., & Edenbrandt, L. (1998). Intelligent computer reporting 'lack of experience': a confidence measure for decision support systems. *Clinical Physiology*, *18*(2), 139–147.

Melluish, T., Saunders, C., Nouretdinov, I., & Vovk, V. (2001). Comparing the Bayes and Typicalness frameworks. In *Proceedings of the 12th European Conference on Machine Learning (ECML'01)*, Vol. 2167 of *Lecture Notes in Computer Science*, pp. 360–371. Springer.

Nouretdinov, I., Melluish, T., & Vovk, V. (2001a). Ridge regression confidence machine. In *Proceedings of the 18th International Conference on Machine Learning (ICML'01)*, pp. 385–392, San Francisco, CA. Morgan Kaufmann.

Nouretdinov, I., Vovk, V., Vyugin, M. V., & Gammerman, A. (2001b). Pattern recognition and density estimation under the general i.i.d. assumption. In *Proceedings of the 14th Annual Conference on Computational Learning Theory and 5th European Conference on Computational Learning Theory*, Vol. 2111 of *Lecture Notes in Computer Science*, pp. 337–353. Springer.

Papadopoulos, H. (2008). Inductive Conformal Prediction: Theory and application to neural networks. In Fritzsche, P. (Ed.), *Tools in Artificial Intelligence*, chap. 18, pp. 315–330. InTech, Vienna, Austria. URL http://www.intechopen.com/download/pdf/pdfs_id/5294.

Papadopoulos, H., Gammerman, A., & Vovk, V. (2008). Normalized nonconformity measures for regression conformal prediction. In *Proceedings of the IASTED International Conference on Artificial Intelligence and Applications (AIA 2008)*, pp. 64–69. ACTA Press.

Papadopoulos, H., Gammerman, A., & Vovk, V. (2009a). Confidence predictions for the diagnosis of acute abdominal pain. In Iliadis, L., Vlahavas, I., & Bramer, M. (Eds.), *Artificial Intelligence Applications & Innovations III*, Vol. 296 of *IFIP International Federation for Information Processing*, pp. 175–184. Springer.

Papadopoulos, H., Papatheocharous, E., & Andreou, A. S. (2009b). Reliable confidence intervals for software effort estimation. In *Proceedings of the 2nd Workshop on Artificial Intelligence Techniques in Software Engineering (AISEW 2009)*, Vol. 475 of *CEUR Workshop Proceedings*. CEUR-WS.org. URL http://ceur-ws.org/Vol-475/AISEW2009/22-pp-211-220-208.pdf.

Papadopoulos, H., Proedrou, K., Vovk, V., & Gammerman, A. (2002a). Inductive confidence machines for regression. In *Proceedings of the 13th European Conference on Machine Learning (ECML'02)*, Vol. 2430 of *Lecture Notes in Computer Science*, pp. 345–356. Springer.

Papadopoulos, H., Vovk, V., & Gammerman, A. (2002b). Qualified predictions for large data sets in the case of pattern recognition. In *Proceedings of the 2002 International Conference on Machine Learning and Applications (ICMLA'02)*, pp. 159–163. CSREA Press.

Papadopoulos, H., Vovk, V., & Gammerman, A. (2007). Conformal prediction with neural networks. In *Proceedings of the 19th IEEE International Conference on Tools with Artificial Intelligence (ICTAI'07)*, Vol. 2, pp. 388–395. IEEE Computer Society.

Proedrou, K., Nouretdinov, I., Vovk, V., & Gammerman, A. (2002). Transductive confidence machines for pattern recognition. In *Proceedings of the 13th European Conference on Machine Learning (ECML'02)*, Vol. 2430 of *Lecture Notes in Computer Science*, pp. 381–390. Springer.

Rasmussen, C. E., Neal, R. M., Hinton, G. E., Van Camp, D., Revow, M., Ghahramani, Z., Kustra, R., & Tibshirani, R. (1996). DELVE: Data for evaluating learning in valid experiments. URL http://www.cs.toronto.edu/~delve/.

Rasmussen, C. E., & Williams, C. K. I. (2006). *Gaussian Processes for Machine Learning*. MIT Press.

Saunders, C., Gammerman, A., & Vovk, V. (1999). Transduction with confidence and credibility. In *Proceedings of the 16th International Joint Conference on Artificial Intelligence*, Vol. 2, pp. 722–726, Los Altos, CA. Morgan Kaufmann.

Saunders, C., Gammerman, A., & Vovk, V. (2000). Computationally efficient transductive machines. In *Proceedings of the Eleventh International Conference on Algorithmic Learning Theory (ALT'00)*, Vol. 1968 of *Lecture Notes in Artificial Intelligence*, pp. 325–333, Berlin. Springer.

Shahmuradov, I. A., Solovyev, V. V., & Gammerman, A. J. (2005). Plant promoter prediction with confidence estimation. *Nucleic Acids Research*, *33*(3), 1069–1076.

Valiant, L. G. (1984). A theory of the learnable. *Communications of the ACM*, *27*(11), 1134–1142.

Vovk, V., Gammerman, A., & Shafer, G. (2005). *Algorithmic Learning in a Random World*. Springer, New York.

Zhang, J., Li, G., Hu, M., Li, J., & Luo, Z. (2008). Recognition of hypoxia EEG with a preset confidence level based on EEG analysis. In *Proceedings of the International Joint Conference on Neural Networks (IJCNN 2008), part of the IEEE World Congress on Computational Intelligence (WCCI 2008)*, pp. 3005–3008. IEEE.